\documentclass{article}

\usepackage{PRIMEarxiv}

\usepackage[utf8]{inputenc} 
\usepackage[T1]{fontenc}    
\usepackage{hyperref}       
\usepackage{url}            
\usepackage{booktabs}       
\usepackage{amsfonts}       
\usepackage{nicefrac}       
\usepackage{microtype}      
\usepackage{lipsum}
\usepackage{fancyhdr}       
\usepackage{graphicx}       
\graphicspath{{media/}}     

\title{The Describe-Then-Generate Bottleneck: How VLM Descriptions Alter Image Generation Outcomes
	\thanks{\textit{\underline{Sports Vision Research}}} 
}
\author{
	Sai Varun, Kodathala\thanks{Corresponding author. Email: varun@sportsvision.ai} \\
	Research and Development \\
	Sports Vision, Inc. \\
	Minnetonka, USA\\
	\texttt{varun@sportsvision.ai}
	\And
	Rakesh, Vunnam \\
	Research and Development \\
	Vizworld, Inc. \\
	Minnetonka, USA\\
	\texttt{rakesh@vizworld.ai} \\
}

\begin{document}
\maketitle

\begin{abstract}
With the increasing integration of multimodal AI systems in creative workflows, understanding information loss in vision-language-vision pipelines has become important for evaluating system limitations. However, the degradation that occurs when visual content passes through textual intermediation remains poorly quantified. In this work, we provide empirical analysis of the describe-then-generate bottleneck, where natural language serves as an intermediate representation for visual information. We generated 150 image pairs through the describe-then-generate pipeline and applied existing metrics (LPIPS, SSIM, and color distance) to measure information preservation across perceptual, structural, and chromatic dimensions. Our evaluation reveals that 99.3\% of samples exhibit substantial perceptual degradation and 91.5\% demonstrate significant structural information loss, providing empirical evidence that the describe-then-generate bottleneck represents a measurable and consistent limitation in contemporary multimodal systems.

\end{abstract}


\section{Introduction}

Recent advances in text-to-image generative models have achieved remarkable capabilities in synthesizing photorealistic imagery from natural language descriptions. Systems such as Stable Diffusion \cite{rombach2022high} and DALL-E 2 \cite{ramesh2022hierarchical} demonstrate unprecedented performance in translating textual prompts into high-fidelity visual content, fundamentally transforming creative workflows and democratizing image generation capabilities. However, the proliferation of multimodal AI systems has revealed a critical architectural limitation in workflows that require visual content to pass through linguistic intermediation - a phenomenon we term the "Describe-then-Generate" bottleneck.

This bottleneck manifests in any pipeline where visual information undergoes textual translation before regeneration: Original Image → Vision-Language Model (VLM) Description → Generated Image, or User Intent → VLM Description → Generated Image. The fundamental challenge lies in the inherently lossy nature of natural language as a representation medium for high-dimensional visual information. Images contain rich, continuous data encompassing precise color gradients, subtle textural variations, complex spatial relationships, and nuanced expressions that resist complete encapsulation within the discrete token space of textual descriptions.

\begin{figure}[h]
	\centering
	\includegraphics[width=0.9\textwidth]{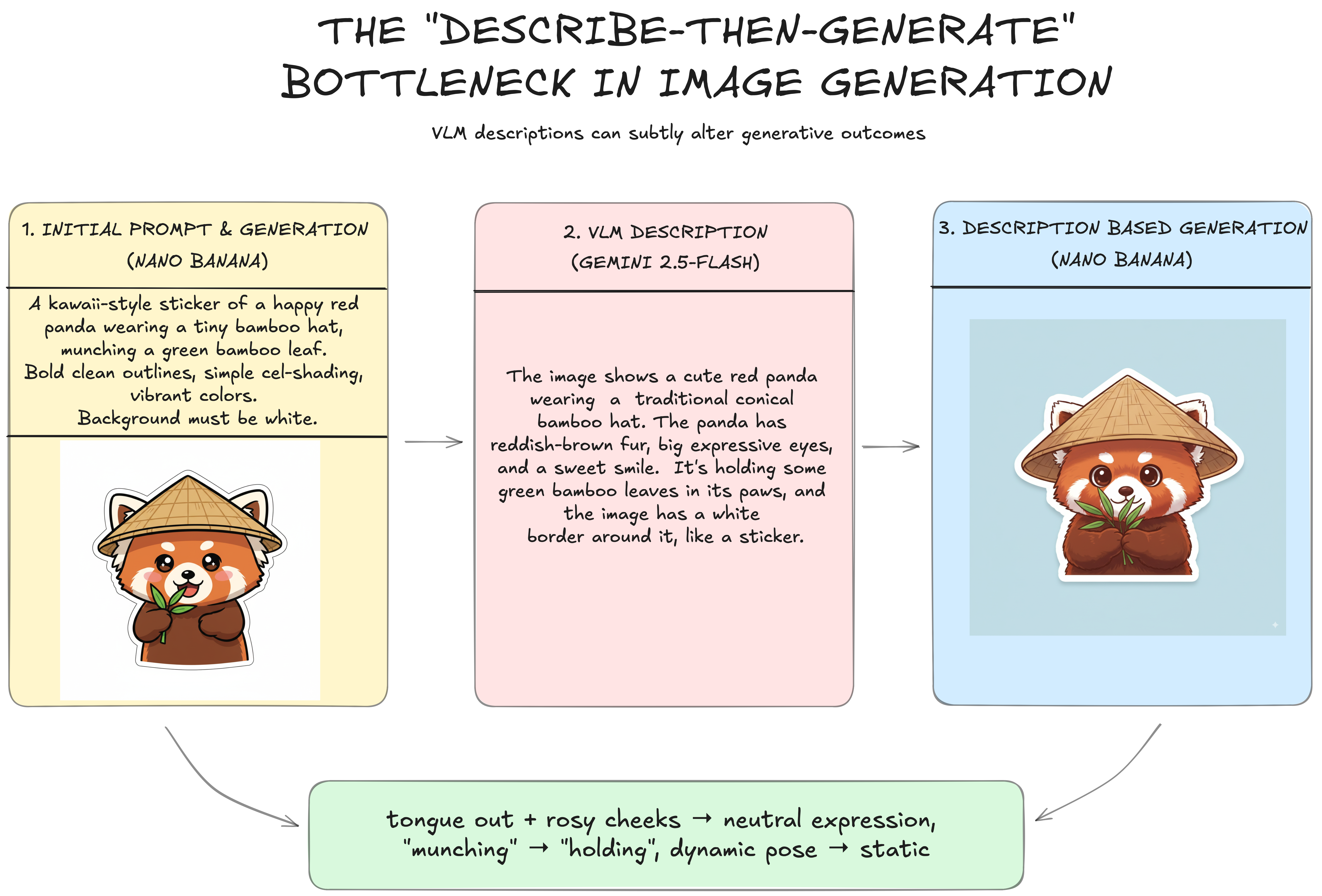}
	\caption{Conceptual illustration of the describe-then-generate bottleneck showing information loss through VLM intermediation. The diagram demonstrates how specific visual details (tongue position, facial expression, pose dynamics) are lost or altered when transitioning from direct prompt generation through textual description to final regeneration.}
	\label{fig:bottleneck_concept}
\end{figure}

This bottleneck extends beyond theoretical interest to impact practical applications requiring high-fidelity visual reproduction. Generative model evaluation becomes problematic when the stochastic nature of diffusion models produces outputs with varying degrees of semantic fidelity for identical prompts, necessitating objective methods for measuring information preservation. Cross-model comparison is complicated when different generative architectures yield outputs with varying degrees of semantic consistency for equivalent textual inputs. Dataset augmentation in domains with limited authentic imagery, such as medical imaging, astronomical observations, or cultural heritage preservation, requires rigorous evaluation of semantic preservation to ensure synthetic data reliability.

The information degradation manifests practically through several observable phenomena: style drift, where detailed artistic techniques become reduced to generic descriptors; detail loss, where intricate patterns are abstracted to vague terms; atmospheric degradation, where subtle mood and lighting conditions disappear entirely; compositional approximation, where precise spatial relationships become imprecise textual descriptions; and chromatic desaturation, where specific color palettes are reduced to basic color terminology.

The mechanisms underlying this information degradation operate across multiple dimensions. Semantic compression occurs when VLMs encode complex visual scenes into finite textual vocabularies, inevitably losing continuous visual properties such as exact colorimetric values, precise geometric arrangements, and subtle stylistic nuances. Attention bias in vision-language architectures typically prioritizes semantically salient objects while neglecting background details, illumination characteristics, and compositional subtleties that contribute significantly to overall visual aesthetics. Linguistic limitations further compound this problem, as natural language lacks sufficient expressive capacity for many visual concepts. The precise articulation of specific color gradations, artistic brushstroke patterns, or atmospheric qualities remains fundamentally challenging through textual description. Additionally, training and cultural bias in VLMs may cause systematic interpretation of visual content through dataset-specific lenses, potentially omitting culturally-specific visual elements or artistic conventions.

Understanding this bottleneck is crucial for advancing multimodal AI architectures beyond current limitations. It highlights the fundamental challenge of employing natural language as an intermediate representation for visual information, suggesting the necessity for more direct multimodal approaches or enhanced description strategies that better preserve visual semantics across modalities.

This work provides the first systematic characterization of the describe-then-generate bottleneck as a fundamental information-theoretic limitation in contemporary multimodal AI systems. Our analysis establishes this bottleneck as a concrete manifestation of the lossy nature of cross-modal translation, offering insights essential for developing more robust visual-linguistic interfaces and informing the design of next-generation multimodal architectures.

\section{Related Work}
\label{sec:related_work}
The describe-then-generate bottleneck intersects multiple established research domains in computer vision, natural language processing, and multimodal learning. We organize the related work into key areas that collectively frame our theoretical and empirical contributions.

\subsection{Multimodal Information Theory and Bottleneck Analysis}
Information-theoretic approaches to multimodal learning provide the theoretical foundation for understanding cross-modal information loss. The Information Bottleneck Principle \cite{tishby1999information} establishes fundamental limits on information transmissionion through intermediate representations. Wu et al. \cite{wu2025learning} proposed Optimal Multimodal Information Bottleneck (OMIB) frameworks that address imbalanced task-relevant information across modalities through theoretically-derived regularization bounds, demonstrating significant improvements in downstream task performance. Almudévar et al. \cite{almudevar2025aligning} revealed that contrastive losses in multimodal representation learning often fail to remove modality-specific information, leading to systematic misalignment between visual and textual representationsons. Li et al. \cite{li2025lost} quantified visual information loss in vision-language models through embedding reconstruction analysis and nearest-neighbor similarity preservation, demonstrating 40-60\% divergence in visual semantic relationships after textual intermediation.

\subsubsection{Vision-Language Alignment and Semantic Preservation}
CLIP-style contrastive learning \cite{radford2021learning} has emerged as the dominant paradigm for aligning visual and textual representations, yet recent analyses reveal systematic limitations in preserving fine-grained visual information. Fang et al. \cite{fang2023dynamic} introduced Dynamic Multimodal Information Bottleneck frameworks that explicitly address redundancy and noise in multimodal features while maintaining task-relevant information through sufficiency constraints. Zhang et al. \cite{zhang2025towards} demonstrated that variational information bottleneck techniques can improve visual authentication tasks by enforcing task-specific feature learning in vision-language models, though at the cost of general-purpose visual understanding. The semantic compression challenge extends to temporal modalities, where Tian et al. \cite{tian2024free} developed prompt-based semantic alignment layers to flexibly align compressed video semantics with multiple vision foundation models.

\subsubsection{Cross-Modal Translation and Information Preservation}
The theoretical limits of cross-modal information transmission have been characterized through mutual information metrics and rate-distortion analysis. Peng et al. \cite{peng2024information} conducted systematic information-theoretic analysis of multimodal image translation, demonstrating that while different modalities contain significant mutual information, translated images systematically fail to preserve complete information from source images. This finding directly validates our characterization of the describe-then-generate bottleneck as a fundamental limitation rather than an implementation artifact. Lin et al. \cite{lin2023relaxing} proposed relaxed contrastive objectives that reduce over-penalization of negative pairs in cross-modal learning, though these approaches still rely on textual intermediation that inherently limits visual information transmission.

\subsubsection{Perceptual Losses and Visual Fidelity Metrics}
Perceptual loss functions have been developed to capture visual similarities aligned with human perception rather than pixel-wise differences. Johnson et al. \cite{johnson2016perceptual} demonstrated that perceptual losses based on pre-trained convolutional networks significantly improve style transfer and super-resolution tasks by preserving high-level semantic content. However, these approaches still rely on learned visual representations that may not capture the nuanced visual information lost during textual description. Amir et al. \cite{amir2021understanding} conducted comprehensive analysis of perceptual distance metrics, revealing systematic biases in how different visual features are weighted in learned representations. Pihlgren et al. \cite{pihlgren2020improving} showed that autoencoders trained with perceptual loss generate embeddings enabling more accurate downstream predictions, particularly for fine-grained visual features that resist textual description. Zhang et al. \cite{zhang2018unreasonable} demonstrated the effectiveness of learned perceptual metrics in capturing human visual similarity judgments, while Wang et al. \cite{wang2004image} provided foundational work on structural similarity measures that complement perceptual approaches.

\subsubsection{Cultural and Training Bias in Multimodal Models}
The describe-then-generate bottleneck is compounded by systematic biases in training data and architectural assumptions. Ananthram et al. \cite{ananthram2024see} characterized Western cultural bias in vision-language models, demonstrating that cultural context significantly affects visual interpretation, with models exhibiting systematically better performance on Western visual concepts compared to East Asian culturalural contexts. Ruggeri and Nozza \cite{ruggeri2023multi} conducted multi-dimensional bias analysis in vision-language models, revealing that pre-trained models complete neutral visual descriptions with biased associations particularly affecting representations of female and minority subjects. These biases compound the fundamental information loss problem by filtering visual content through culturally-specific interpretation frameworks that may systematically omit or misrepresent visual elements from underrepresented cultural contexts. Kadiyala et al. \cite{kadiyala2025uncovering} further explored cultural representation disparities, highlighting the need for more inclusive training data and evaluation metrics.

\subsubsection{Concept Bottleneck Models and Interpretable Representations}
Recent work on concept bottleneck models has explored interpretable intermediate representations that factor visual decisions into human-readable concepts. Yang et al. \cite{yang2023language} introduced language-guided concept bottlenecks that leverage large language models to generate concept vocabularies for visual understanding tasks. However, these approaches still fundamentally rely on textual concept descriptions that may inadequately capture visual nuances essential for high-fidelity reconstruction. Prasse et al. \cite{prasse2024dcbm,prasse2025enhancing} developed data-efficient concept bottleneck models that utilize visual regions rather than textual descriptions, partially addressing the limitations of language-mediated visual understanding while maintaining interpretability constraints. Koh et al. \cite{koh2020concept} provided the foundational framework for concept bottleneck models, establishing the theoretical basis for interpretable machine learning through concept-based intermediate representations.

\subsubsection{Fine-Grained Visual Perception and Abstract Reasoning}
Recent research has highlighted the challenges of fine-grained visual perception in multimodal systems. Yan et al. \cite{yan2025visuriddles} demonstrated that fine-grained perception represents a primary bottleneck for multimodal large language models in abstract visual reasoning tasks, introducing the VisuRiddles benchmark to evaluate these limitations systematically. Zhang et al. \cite{zhang2024cross} analyzed cross-modal information flow in multimodal large language models, revealing systematic information loss patterns that align with our theoretical framework. Yao et al. \cite{yao2024deco} proposed methods to decouple token compression from semantic abstraction, addressing some of the fundamental compression challenges we identify. Zhao et al. \cite{zhao2024looking} explored techniques for reducing language bias in large vision-language models through multimodal dual-attention mechanisms, while Jin et al. \cite{jin2025multimodal} applied multimodal conditional information bottleneck frameworks to AI-generated image detection tasks.

\subsubsection{Semantic Compression and Context Extension}
The challenge of semantic compression in multimodal systems has been addressed through various approaches. Fei et al. \cite{fei2024extending} demonstrated semantic compression techniques for extending context windows in large language models, though these methods face similar trade-offs between compression and information preservation that characterize our bottleneck. Tian et al. \cite{tian2025identifying} identified and mitigated position bias in multi-image vision-language models, revealing systematic biases that compound information loss during visual processing. Ismail et al. \cite{ismail2023right} explored distribution-sensitive losses for improving semantic consistency in text-to-image generation, addressing some of the quality degradation we observe in describe-then-generate pipelines.

\subsubsection{Geometric and Contrastive Multimodal Learning}
Geometric approaches to multimodal representation learning offer alternative perspectives on cross-modal alignment. Poklukar et al. \cite{poklukar2022geometric} developed geometric multimodal contrastive representation learning methods that preserve structural relationships across modalities, though these still face fundamental limitations when visual information must pass through textual intermediation. Ma et al. \cite{ma2023ccim} proposed cross-modal cross-lingual interactive image translation frameworks, while Liu et al. \cite{liu2025principled} established principled foundations for multimodal representation learning that inform our theoretical analysis.

Prior work has addressed individual components of cross-modal information loss, vision-language alignment, perceptual metrics, and prompt optimization, but has not systematically characterized the describe-then-generate bottleneck as a fundamental architectural limitation in multimodal AI systems. Our work provides the first comprehensive theoretical and empirical framework for understanding how textual intermediation systematically constrains visual information preservation across different generative architectures and application domains.

\section{Dataset Construction for Bottleneck Analysis}
\label{dataset}
To investigate the describe-then-generate bottleneck, we constructed a specialized dataset that captures the complete information flow through vision-language-vision pipelines. Our dataset, \textit{Describe-then-Generate-Bottleneck}, contains 150 randomly curated samples that demonstrate the systematic information loss occurring when visual content passes through textual intermediation.

\subsection{Dataset Methodology}

The dataset construction follows a rigorous four-stage pipeline designed to isolate and quantify the bottleneck effect:

\textbf{Stage 1: Source Selection} - We randomly selected 150 diverse prompts from the zai-org/ImageRewardDB dataset, ensuring broad coverage of visual concepts, artistic styles, and complexity levels.

\textbf{Stage 2: Direct Generation} - Each selected prompt was used to generate an image directly using Gemini-2.5-Flash-Image-Preview, creating our baseline images ($I_1$).

\textbf{Stage 3: Vision-Language Description} - The generated images were processed through Gemini-2.5-Flash VLM using the simple prompt "Describe the Image" to produce detailed textual descriptions.

\textbf{Stage 4: Bottleneck Generation} - The VLM descriptions were used to generate new images ($I_2$) using the same Gemini-2.5-Flash-Image-Preview model, completing the describe-then-generate cycle.

This methodology enables direct comparison between images generated from identical conceptual content but different representational pathways: direct textual prompts versus VLM-mediated descriptions.

\subsection{Dataset Examples}

To illustrate the bottleneck phenomenon, we present three representative examples from our dataset:

\subsubsection{Example 1: Cross-Media Character Fusion}

\textbf{Original Prompt:} "john paul ii in breaking bad with walter white"

\begin{figure}[h]
	\centering
	\includegraphics[width=0.45\textwidth]{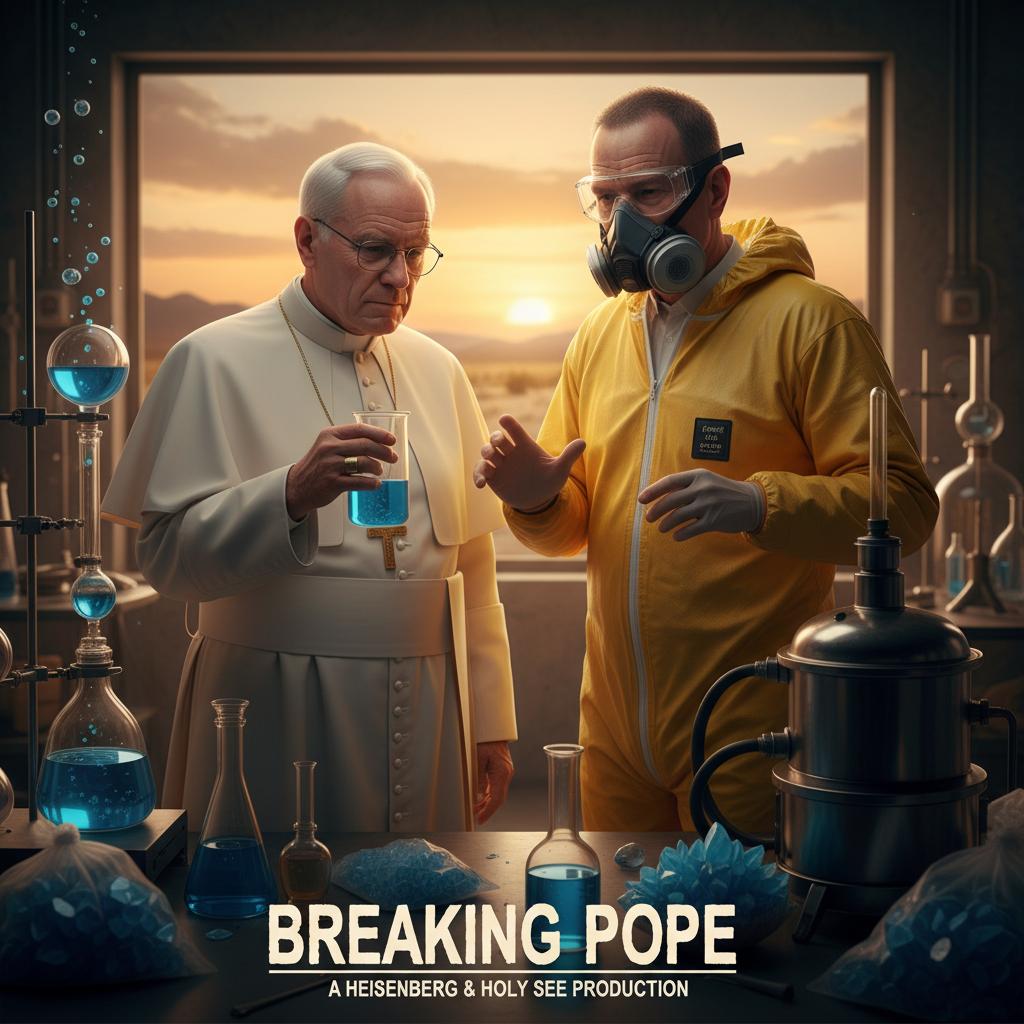}
	\hfill
	\includegraphics[width=0.45\textwidth]{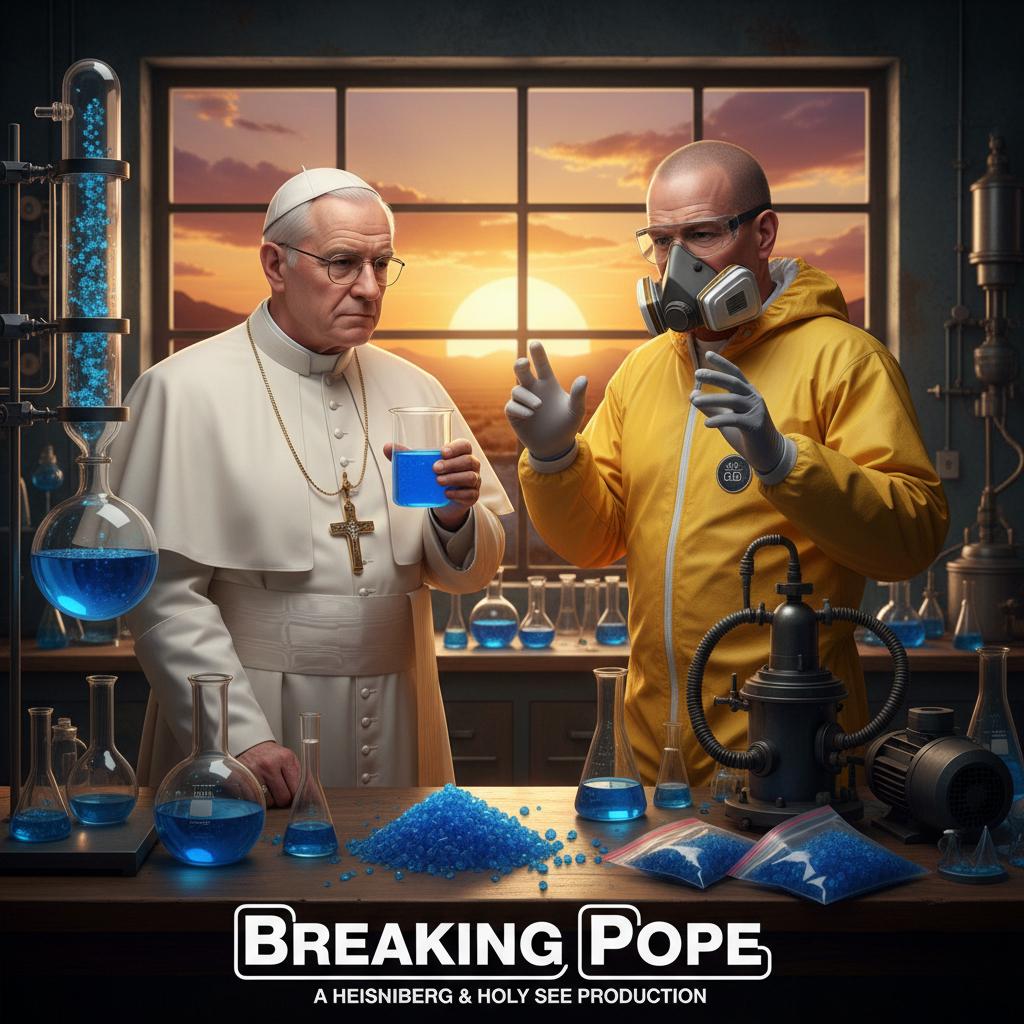}
	\caption{Left: Direct generation from original prompt. Right: Generation from VLM description. Cross-media character fusion showing bottleneck effect on complex character interactions and laboratory setting details.}
	\label{fig:example1}
\end{figure}

\textbf{VLM Description:} This image is a detailed, high-quality digital illustration depicting a humorous and unexpected crossover between the popular TV series 'Breaking Bad' and the Catholic Church. In the foreground, two central figures are positioned in what appears to be a sophisticated, if somewhat makeshift, chemical laboratory. On the left, an elderly man, strongly resembling a Pope (possibly Pope John Paul II), is dressed in traditional pristine white papal attire, including a cassock, a zucchetto (skullcap), and a pectoral cross on a gold chain around his neck. He holds a clear glass beaker containing a vibrant blue liquid, looking at it with a contemplative, perhaps slightly concerned or intrigued expression. His gaze is directed downwards towards the beaker and slightly towards the other man. On the right, a man resembling Walter White (Heisenberg) from 'Breaking Bad' stands in a bright yellow hazmat suit, complete with clear safety goggles and a dark grey respirator mask covering his nose and mouth. He wears white gloves and has his hands raised and gesturing, as if explaining a chemical process or presenting a product. The laboratory setting is rich with details: On the table between the two figures, there are various pieces of scientific glassware, including flasks, beakers, and tubes, many containing the same distinctive blue liquid. Piles of crystalline blue material, strongly reminiscent of 'blue meth' from 'Breaking Bad,' are scattered on the table and contained in clear plastic bags. To the far left, an elaborate glass distillation apparatus features bubbling blue liquid. To the far right, a dark, metallic, multi-tiered chemical processing unit is visible. In the background, a large window reveals a dramatic desert landscape at sunset or sunrise, with a warm palette of orange, yellow, and soft grey in the sky and distant mountains. This vista clearly evokes the setting of 'Breaking Bad' in New Mexico. Prominently displayed at the bottom center of the image in bold white sans-serif font is the title: 'BREAKING POPE.' Below it, in smaller text, reads: 'A HEISENBERG \& HOLY SEE PRODUCTION.' The overall aesthetic is realistic and well-rendered, with particular attention to lighting and textures, creating a striking and amusing juxtaposition of sacred imagery with the gritty world of illicit chemistry.
\subsubsection{Example 2: Architectural Concept Art}

\textbf{Original Prompt:} "mediterranean nuragic shrine, highly detailed, digital painting, artstation, concept art, sharp focus, illustration, art by artgerm and greg rutkowski and raphael lacoste and magali villeneuve"

\begin{figure}[h]
	\centering
	\includegraphics[width=0.45\textwidth]{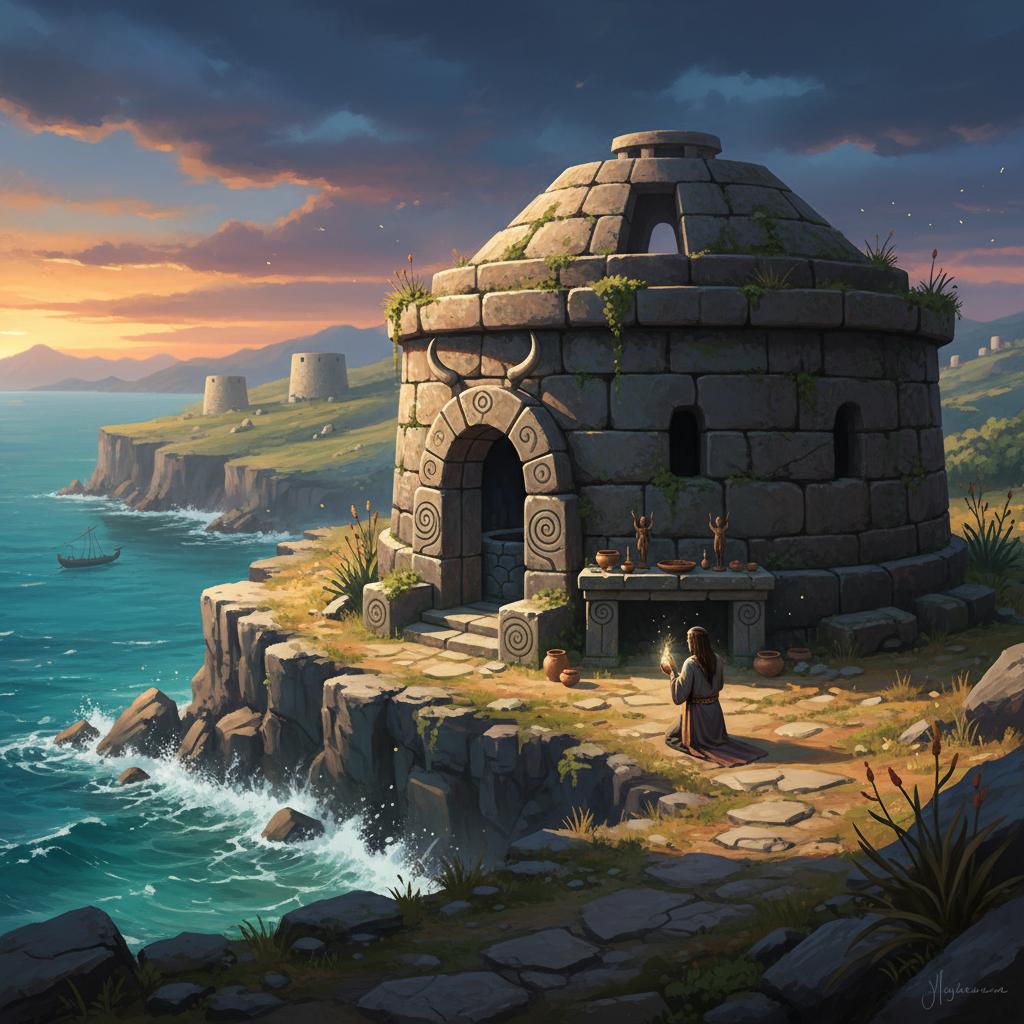}
	\hfill
	\includegraphics[width=0.45\textwidth]{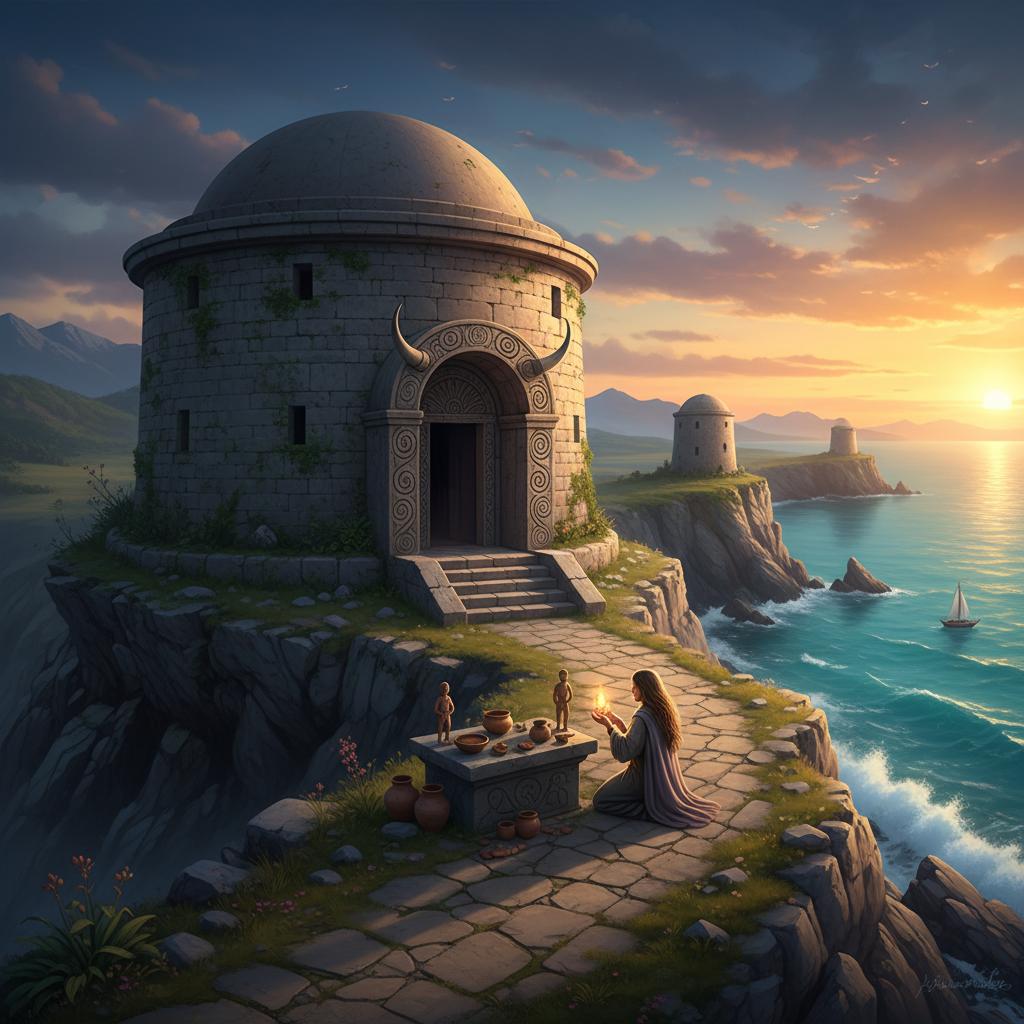}
	\caption{Left: Direct generation from original prompt. Right: Generation from VLM description. Architectural concept art showing loss of artistic style references and technical rendering details.}
	\label{fig:example2}
\end{figure}

\textbf{VLM Description:} This evocative image depicts a serene and mystical coastal scene at what appears to be either dawn or dusk, with warm, golden light painting the sky. In the foreground and midground, a magnificent, ancient stone rotunda or temple stands prominently on a rugged cliff edge. Constructed from large, weathered blocks, the circular building features a domed roof and a prominent arched entrance. Intricate spiral carvings adorn the archway, and a pair of stylized bull horns are placed above the entrance, hinting at its ancient or ritualistic purpose. Small, rectangular openings are visible on its walls, and patches of green vegetation, including vines and grasses, cling to the stone, suggesting its age. Stone steps lead up to the entrance. Immediately in front of the rotunda, a robed figure with long hair is kneeling reverently on a flagstone path. The figure is intensely focused on a small, radiant light source held in their hands, which emits a soft, golden glow. In front of the kneeling figure, an altar-like stone slab holds several small, ancient-looking pottery vessels and two stylized humanoid figurines. Additional pottery items are scattered on the ground around the altar. The rotunda is dramatically perched above a vibrant turquoise sea, where powerful waves crash against the base of the steep, stratified rocky cliffs, creating splashes of white foam. A narrow, well-trodden path of flagstones leads towards the structure. In the distance, further along the rugged coastline, two similar, smaller round stone towers stand on distant headlands, suggesting a larger ancient settlement or series of lookout points. A small, simple sailboat glides peacefully on the calm waters further out. The sky is a dramatic tableau of colors, transitioning from brilliant oranges and yellows near the horizon, where distant mountains are silhouetted, to deeper blues and purples overhead, with wispy clouds catching the last or first light. The overall lighting is soft and warm, casting long shadows and enhancing the ethereal and mystical atmosphere of the scene. The image combines elements of ancient architecture, spiritual ritual, and breathtaking natural beauty, evoking a strong sense of history and wonder.

\subsubsection{Example 3: Fantasy Character Design}

\textbf{Original Prompt:} "innocent-eyed fantasy creature with pure white fur coat, drooping rabbity ears, and cute little fangs. Cartoon character concept."

\begin{figure}[h]
	\centering
	\includegraphics[width=0.45\textwidth]{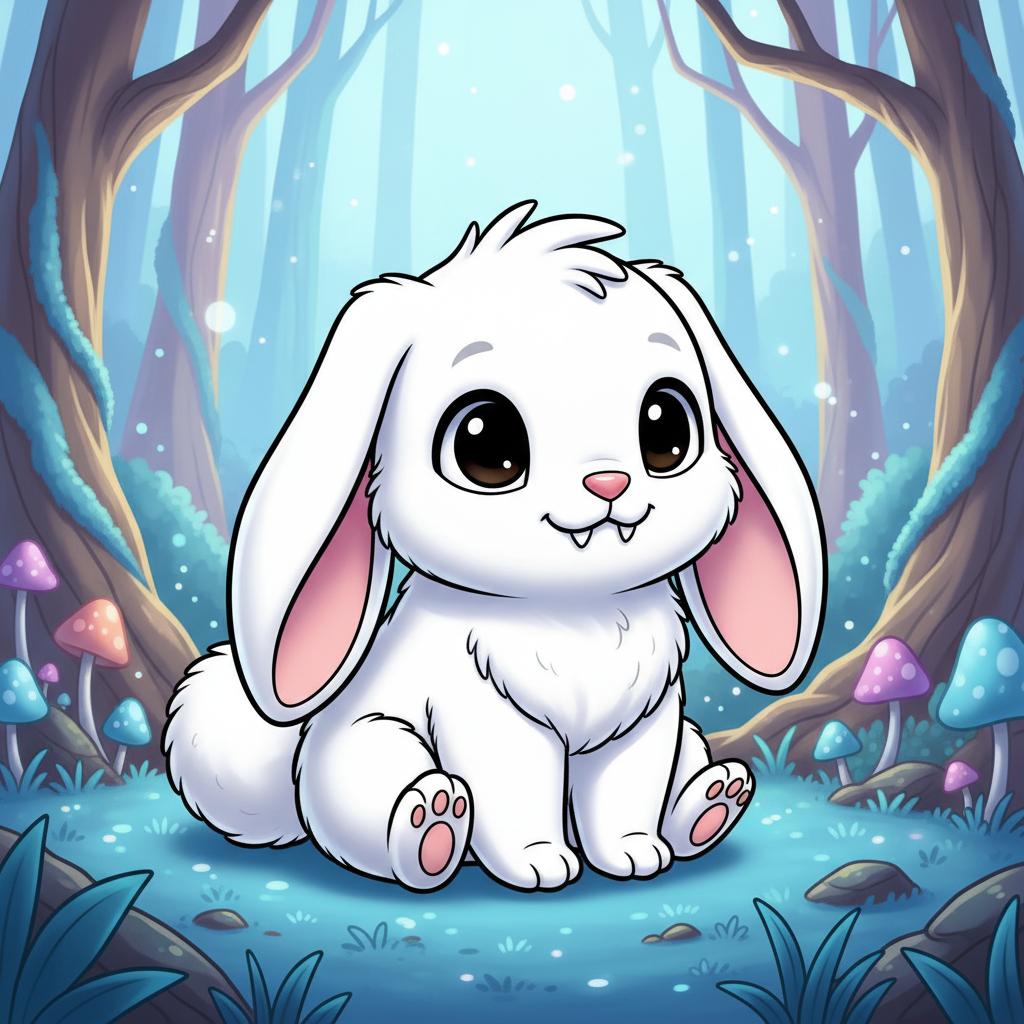}
	\hfill
	\includegraphics[width=0.45\textwidth]{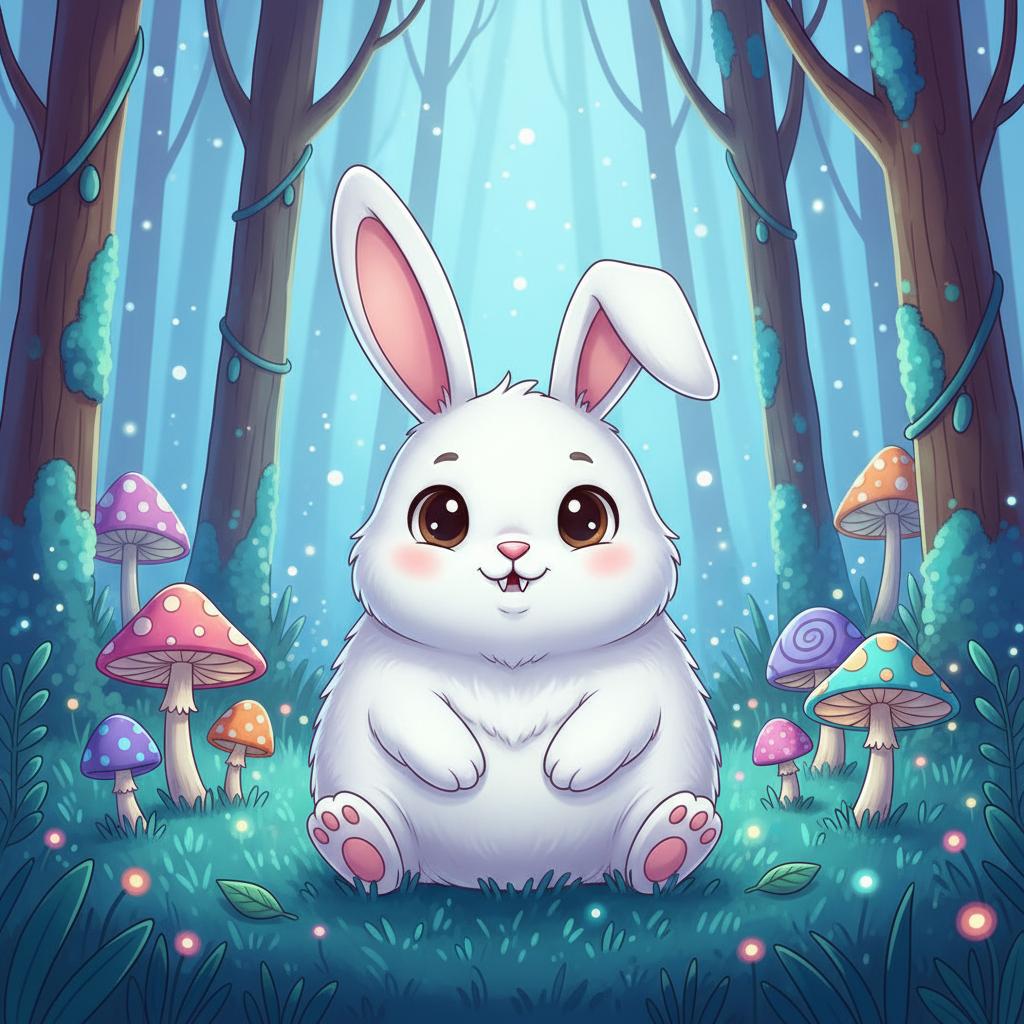}
	\caption{Left: Direct generation from original prompt. Right: Generation from VLM description. Fantasy character design showing environmental expansion vs. character focus in VLM interpretation.}
	\label{fig:example3}
\end{figure}

\textbf{VLM Description:} This is a whimsical and adorable digital illustration featuring a cute white rabbit in an enchanted forest setting. At the center of the image is a charming, plump white rabbit with soft, fluffy fur. It has large, expressive dark brown eyes with prominent white gleams, giving it an innocent and sweet look. Its small pink nose and a subtle, friendly smile reveal two tiny, pointed fangs, giving it a playful, almost 'vampire bunny' appearance rather than menacing. Long, floppy ears with soft pink interiors frame its head. The rabbit is sitting upright, displaying its small pink paw pads on its front feet. The rabbit is situated in a magical forest scene with a predominant cool color palette of blues, greens, and purples. Tall, slender trees with smooth brown trunks rise into a soft blue, ethereal mist in the background. Some of the tree trunks are adorned with patches of bright blue or teal moss or vines. The forest floor is a vibrant teal-blue, covered in short grass, scattered leaves, and small glowing specks or flowers, creating a sparkling effect. Numerous colorful, whimsical mushrooms dot the landscape around the rabbit. These mushrooms come in various sizes and feature shades of pink, purple, light blue, and orange, often with contrasting spots or details on their caps. Tiny white sparkling orbs float through the air, adding to the magical and dreamlike atmosphere. The overall art style is distinctly cartoonish and cute, with clear outlines and smooth digital rendering, reminiscent of a children's storybook illustration or a friendly animated character. The image evokes a sense of wonder, cuteness, and gentle fantasy.

\subsection{Bottleneck Manifestation Analysis}

\begin{figure}[h]
	\centering
	\includegraphics[width=0.45\textwidth]{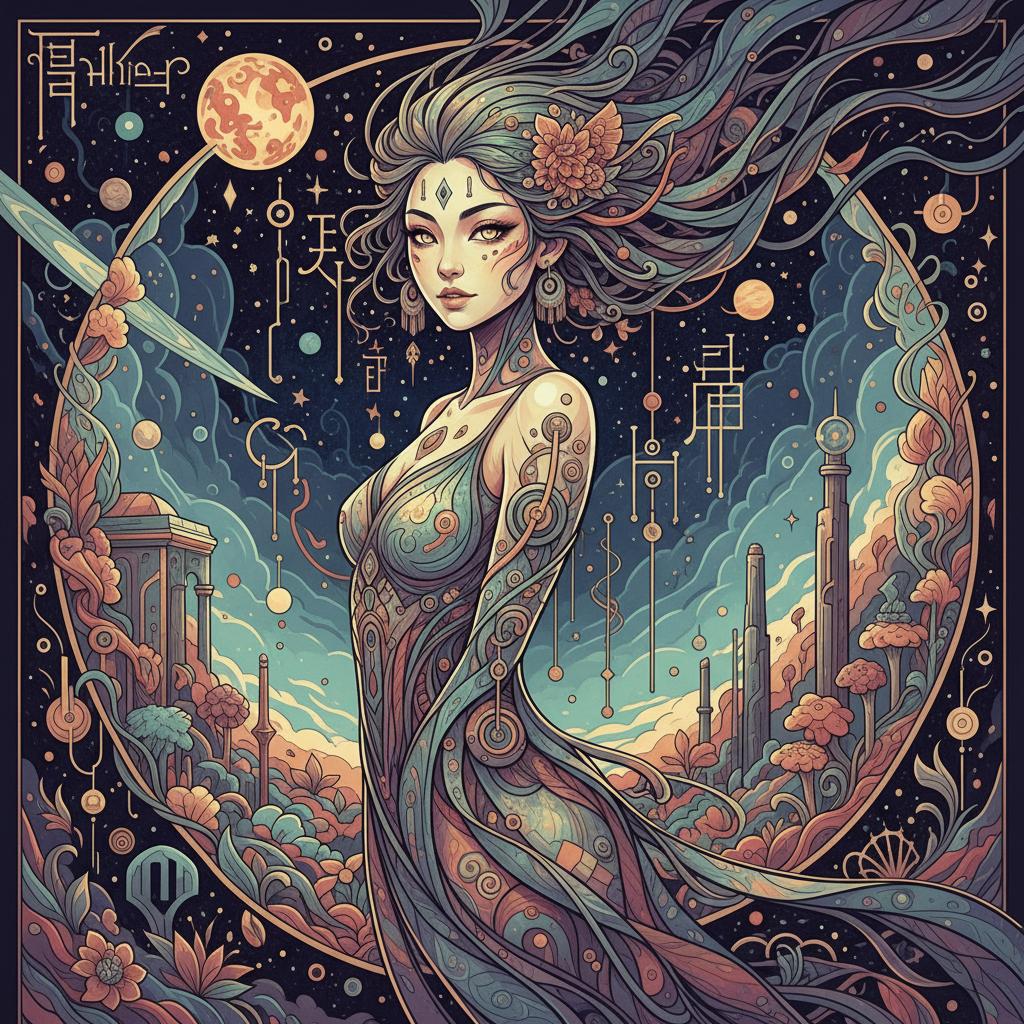}
	\hfill
	\includegraphics[width=0.45\textwidth]{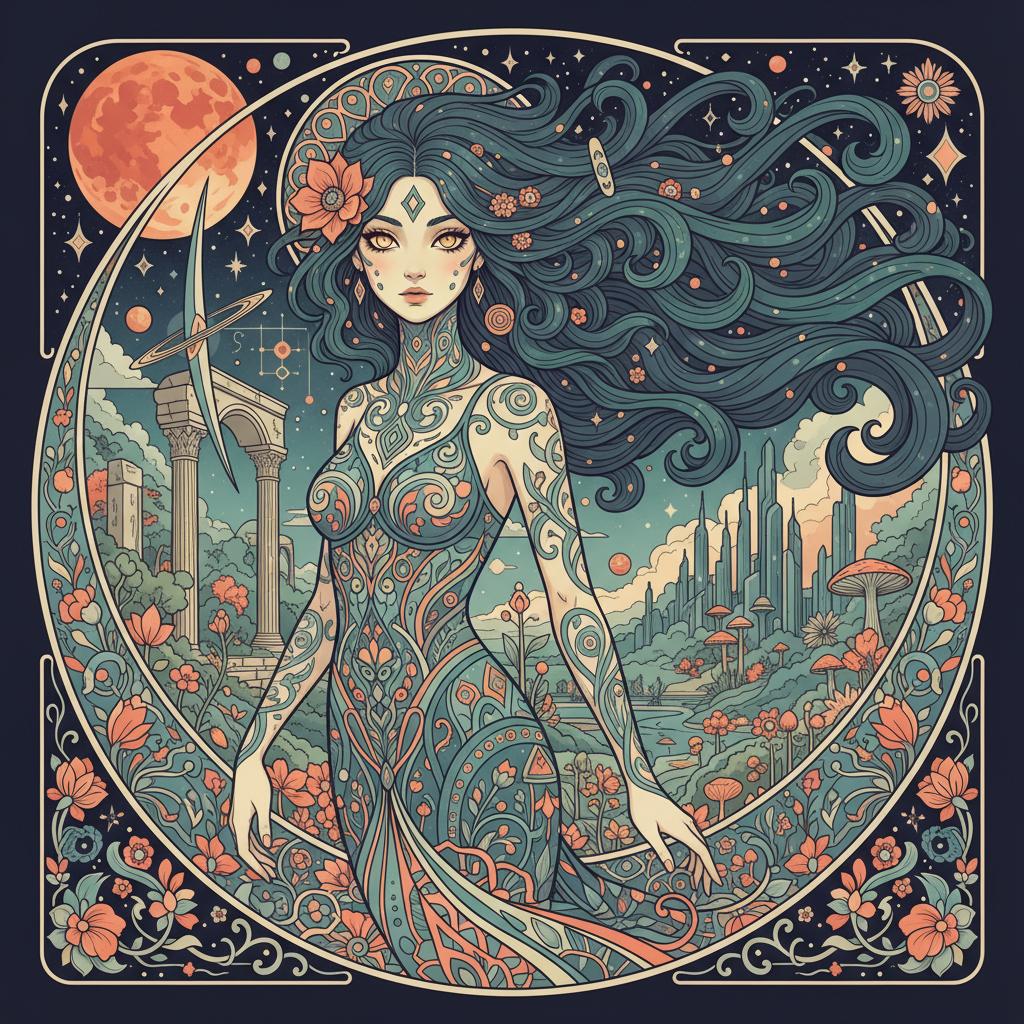}
	\caption{Left: Prompt-generated image. Right: Description-generated image. Example 2 showing detail degradation in artistic elements.}
	\label{fig:comp2}
\end{figure}

\begin{figure}[h]
	\centering
	\includegraphics[width=0.45\textwidth]{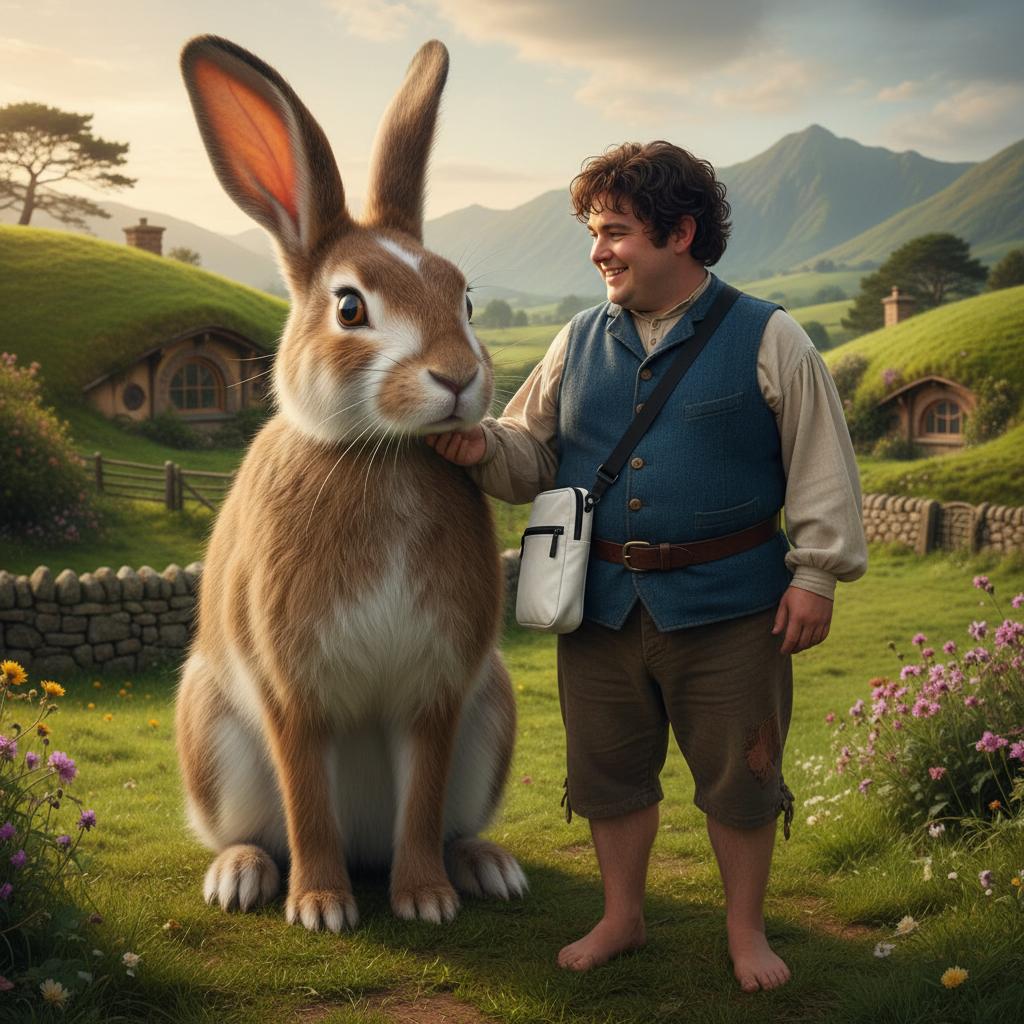}
	\hfill
	\includegraphics[width=0.45\textwidth]{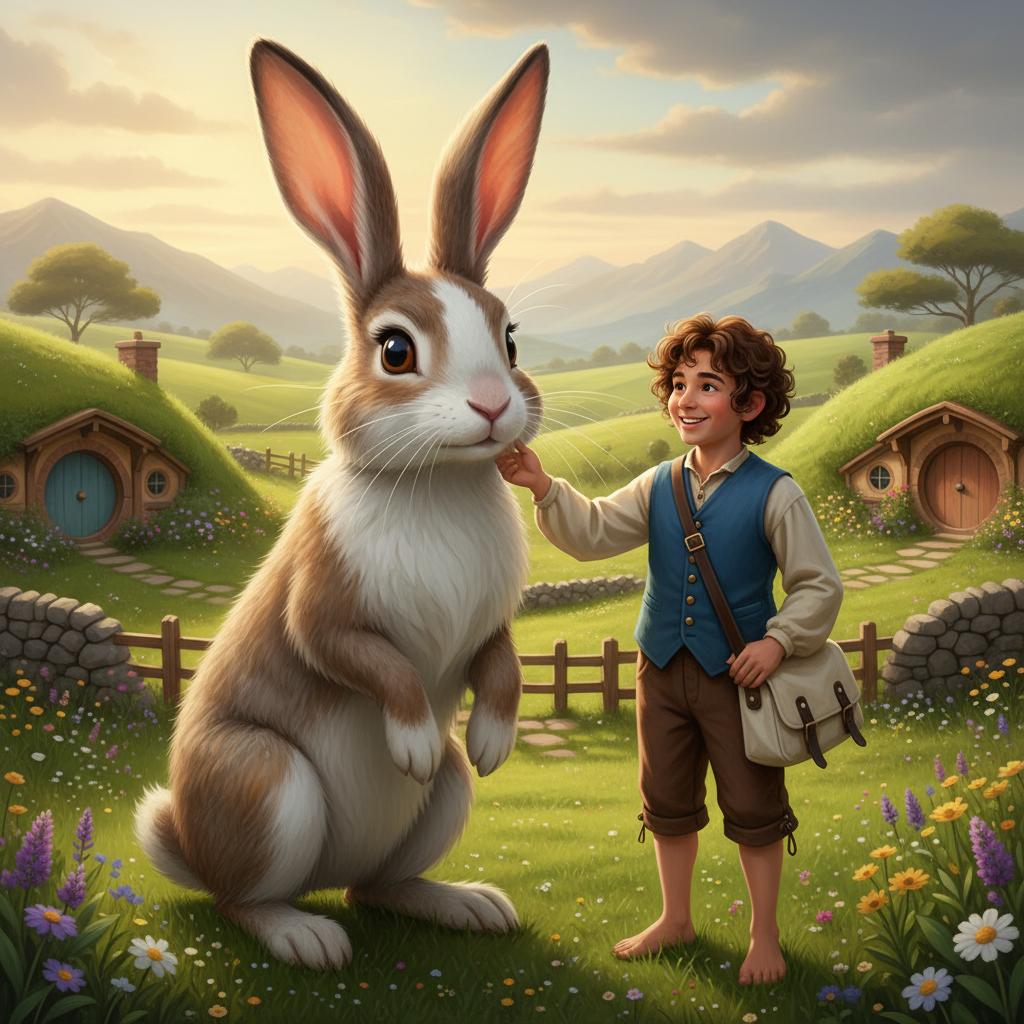}
	\caption{Left: Prompt-generated image. Right: Description-generated image. Example 3 illustrating compositional approximation effects.}
	\label{fig:comp3}
\end{figure}

Figures \ref{fig:comp1}, \ref{fig:comp2}, and \ref{fig:comp3} present additional examples demonstrating various aspects of information loss through the describe-then-generate pipeline:

\begin{figure}[h]
	\centering
	\includegraphics[width=0.45\textwidth]{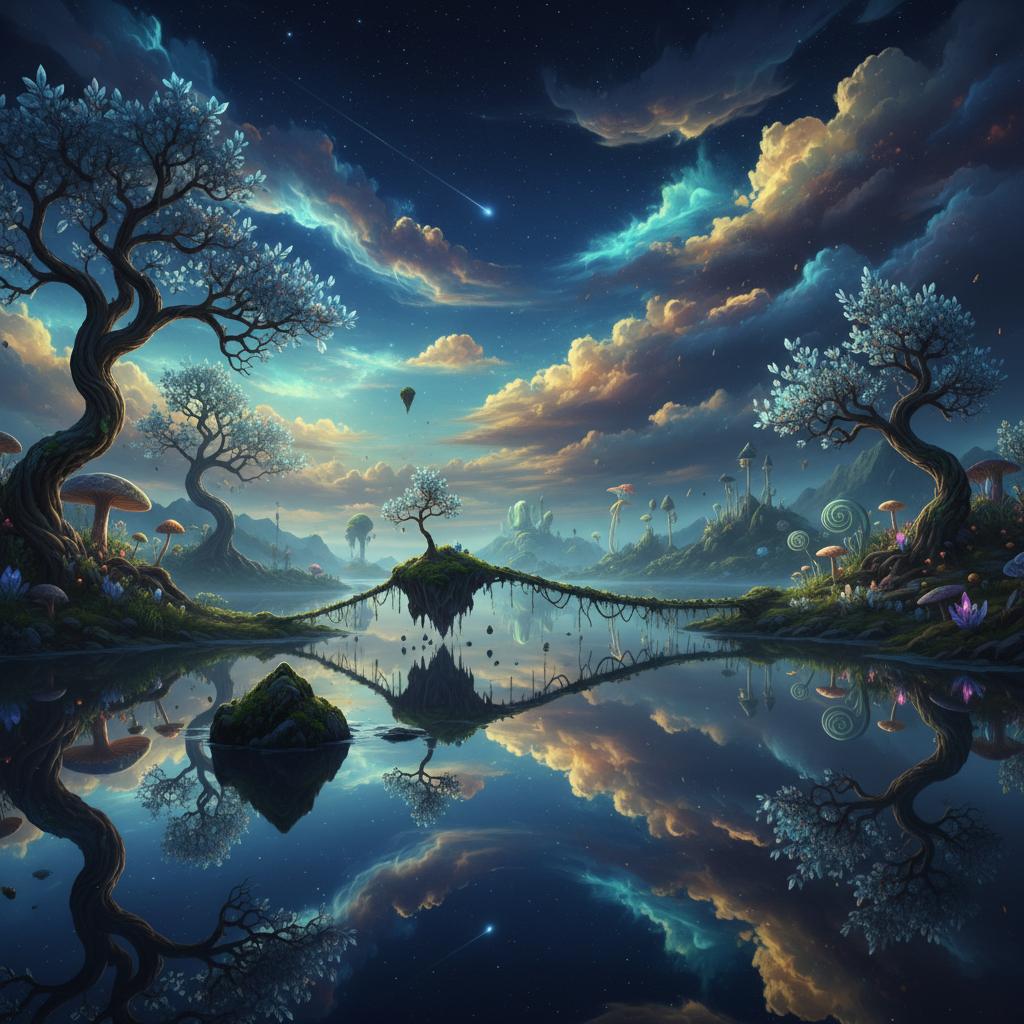}
	\hfill
	\includegraphics[width=0.45\textwidth]{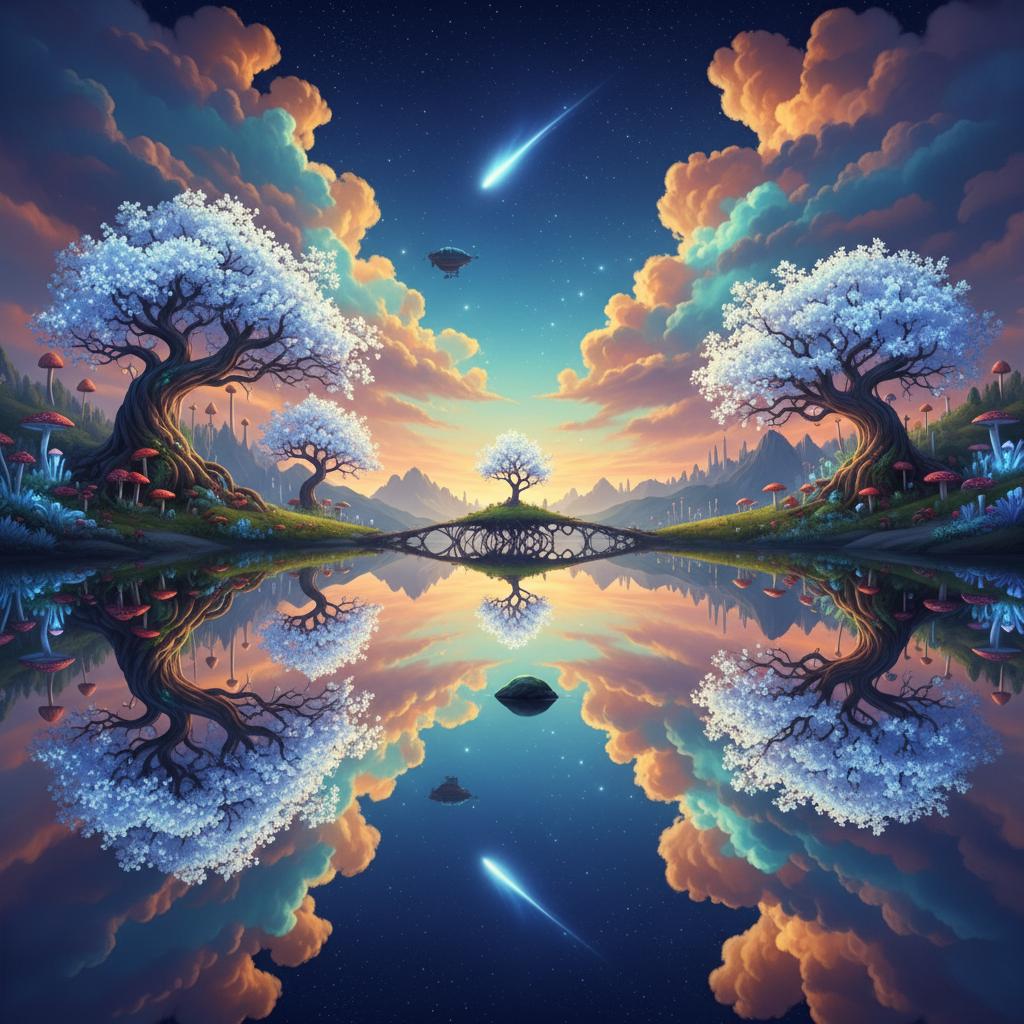}
	\caption{Left: Prompt-generated image. Right: Description-generated image. Example 1 demonstrating style drift in landscape composition.}
	\label{fig:comp1}
\end{figure}

The examples presented demonstrate characteristic bottleneck phenomena including style drift (where artistic techniques become generic descriptors), detail loss (where specific visual elements are abstracted), and compositional approximation (where precise spatial relationships become imprecise textual descriptions). This dataset provides the empirical foundation for systematically characterizing the describe-then-generate bottleneck as a fundamental limitation in contemporary multimodal AI systems.

\section{Evaluation Framework}

To systematically evaluate the describe-then-generate bottleneck, we establish a focused evaluation framework that measures information preservation across three critical dimensions of visual fidelity. In our experimental setup, $I_1$ represents images generated directly from user prompts using Gemini-2.5-Flash-Image-Preview, while $I_2$ represents images generated from VLM descriptions (obtained via Gemini-2.5-Flash) of $I_1$ using the same image generation model as described in Section \ref{dataset}.

\subsection{Metric Selection Rationale}

We selected three complementary metrics based on their demonstrated sensitivity to different aspects of visual information loss in cross-modal translation. Our selection deliberately excludes high-level semantic metrics (such as CLIP similarity) that may remain artificially elevated even when significant fine-grained visual information is lost, potentially masking the bottleneck effect we aim to quantify.

The chosen metrics target distinct but complementary aspects of visual degradation: perceptual features that affect human visual perception, structural patterns that define spatial relationships, and chromatic properties that convey mood and aesthetic qualities. This multi-dimensional approach provides comprehensive coverage of information preservation without redundancy in measurement domains.

\subsection{Visual Similarity Metrics}

\subsubsection{Perceptual Metrics}

\textbf{LPIPS (Learned Perceptual Image Patch Similarity)}

LPIPS leverages VGG16 features to measure perceptual distance aligned with human visual perception \cite{zhang2018unreasonable}. This metric captures mid-level visual features including textures, edges, and spatial arrangements that are particularly vulnerable to loss during textual intermediation:

\begin{equation}
	LPIPS_{distance} = ||F_{VGG16}(I_1) - F_{VGG16}(I_2)||_2
\end{equation}

where $F_{VGG16}$ represents deep features extracted from VGG16 layers. Lower values indicate higher perceptual similarity. LPIPS is specifically sensitive to the types of visual details that resist complete encapsulation in textual descriptions, making it ideal for detecting bottleneck effects.

\subsubsection{Structural Metrics}

\textbf{SSIM (Structural Similarity Index)}

SSIM measures structural similarity by comparing luminance, contrast, and structure between images \cite{wang2004image}. This metric is particularly sensitive to geometric distortions and compositional changes that occur when spatial relationships are translated through textual description:

\begin{equation}
	SSIM = \frac{(2\mu_1\mu_2 + c_1)(2\sigma_{12} + c_2)}{(\mu_1^2 + \mu_2^2 + c_1)(\sigma_1^2 + \sigma_2^2 + c_2)}
\end{equation}

Range: [0, 1], where $\mu_1, \mu_2$ are local means, $\sigma_1^2, \sigma_2^2$ are local variances, $\sigma_{12}$ is local covariance, and $c_1, c_2$ are stabilization constants. SSIM effectively captures structural preservation while being robust to uniform illumination changes.

\subsubsection{Chromatic Metrics}

\textbf{Color Distance}

Color distance quantifies statistical color preservation between images by measuring differences in color distribution across RGB channels:

\begin{equation}
	Color_{distance} = \sqrt{\sum_{c \in \{R,G,B\}} (\mu_{c,1} - \mu_{c,2})^2 + (\sigma_{c,1} - \sigma_{c,2})^2}
\end{equation}

where $\mu_{c,i}$ and $\sigma_{c,i}$ represent the mean and standard deviation of color channel $c$ in image $i$. This metric captures chromatic degradation that occurs when nuanced color relationships are reduced to discrete textual color descriptors during the describe-then-generate process.

\section{Results and Analysis}

Our evaluation of 150 image pairs provides empirical evidence for the describe-then-generate bottleneck across multiple dimensions of visual fidelity. The analysis reveals systematic information loss when visual content undergoes textual intermediation, with particularly pronounced degradation in perceptual and structural preservation.

\subsection{Quantitative Analysis}

Table \ref{tab:metrics_summary} presents the distribution of similarity scores across our three-metric evaluation framework. The results demonstrate substantial information loss across all measured dimensions, with notably severe degradation in structural and perceptual preservation.

\begin{table}[ht]
	\caption{Summary Statistics for Visual Similarity Metrics}
	\centering
	\begin{tabular}{lccccc}
		\toprule
		Metric & Mean & Std & Min & Max & Median \\
		\midrule
		LPIPS Distance & 0.635 & 0.054 & 0.463 & 0.754 & 0.641 \\
		SSIM Score & 0.355 & 0.115 & 0.061 & 0.643 & 0.351 \\
		Color Distance & 30.17 & 18.84 & 3.95 & 120.88 & 23.66 \\
		\bottomrule
	\end{tabular}
	\label{tab:metrics_summary}
\end{table}

The LPIPS distance scores (mean = 0.635, std = 0.054) indicate substantial perceptual degradation, with 99.3\% of samples exceeding the established degradation threshold of 0.5. The relatively low standard deviation suggests consistent rather than random information loss across diverse image content.

SSIM scores (mean = 0.355, std = 0.115) demonstrate significant structural information loss, with 91.5\% of samples falling below the preservation threshold of 0.5. The higher variance indicates differential structural preservation across content types, suggesting that certain visual compositions are more vulnerable to bottleneck effects than others.

Color distance measurements reveal variable chromatic preservation, with a mean distance of 30.17 and high standard deviation of 18.84, indicating content-dependent color degradation patterns during textual intermediation.

\subsection{Bottleneck Evidence Analysis}

Table \ref{tab:bottleneck_evidence} quantifies the prevalence of information degradation across our evaluation framework. The analysis establishes clear thresholds for bottleneck detection based on established perceptual studies and statistical analysis of metric distributions.

\begin{table}[ht]
	\caption{Prevalence of Information Degradation Across Visual Dimensions}
	\centering
	\begin{tabular}{lcc}
		\toprule
		Metric & Degradation Threshold & Samples Exceeding Threshold \\
		\midrule
		LPIPS Distance & $> 0.5$ & 99.3\% \\
		SSIM Score & $< 0.5$ & 91.5\% \\
		Color Distance & $> 44$ & 19.9\% \\
		\bottomrule
	\end{tabular}
	\label{tab:bottleneck_evidence}
\end{table}

The near-universal degradation in perceptual (99.3\%) and structural (91.5\%) dimensions provides strong empirical support for the bottleneck hypothesis. Color preservation shows moderate degradation rates (19.9\%), indicating measurable but less pervasive chromatic information loss during textual intermediation.

\subsection{Inter-Metric Correlation Analysis}

Table \ref{tab:correlations} presents correlation coefficients between our three metrics, revealing the relationship between different dimensions of information loss.

\begin{table}[ht]
	\caption{Inter-Metric Correlation Analysis}
	\centering
	\begin{tabular}{lcc}
		\toprule
		Metric Pair & Correlation Coefficient & Relationship Strength \\
		\midrule
		LPIPS vs SSIM & -0.494 & Moderate \\
		LPIPS vs Color Distance & 0.050 & Weak \\
		SSIM vs Color Distance & 0.017 & Weak \\
		\bottomrule
	\end{tabular}
	\label{tab:correlations}
\end{table}

The moderate negative correlation between LPIPS and SSIM (-0.494) indicates that perceptual and structural degradation are related but measure distinct aspects of information loss. The weak correlations between chromatic and other metrics suggest that color preservation operates relatively independently from perceptual and structural dimensions.

\subsection{Extreme Case Analysis}

Table \ref{tab:worst_cases} presents samples exhibiting the most severe bottleneck effects, characterized by degradation across multiple metric dimensions simultaneously.

\begin{table}[ht]
	\caption{Samples with Most Severe Bottleneck Manifestation}
	\centering
	\begin{tabular}{lccc}
		\toprule
		Sample ID & LPIPS Distance & SSIM Score & Color Distance \\
		\midrule
		003407-0036 & 0.656 & 0.140 & 115.58 \\
		005752-0065 & 0.682 & 0.244 & 120.88 \\
		007726-0040 & 0.659 & 0.072 & 67.62 \\
		005342-0117 & 0.693 & 0.204 & 60.19 \\
		006664-0082 & 0.686 & 0.314 & 69.85 \\
		\bottomrule
	\end{tabular}
	\label{tab:worst_cases}
\end{table}

These sample IDs can be retrieved from the publicly available dataset for detailed analysis.
\begin{center}
	\url{https://huggingface.co/datasets/sportsvision/Describe-then-Generate-Bottleneck}
\end{center}

These extreme cases demonstrate compound information loss, with particularly severe structural degradation (SSIM scores below 0.3) combined with substantial perceptual changes (LPIPS scores above 0.65). Such cases represent the most pronounced manifestations of the bottleneck effect in our dataset.

The weak correlations between different degradation dimensions suggest that the bottleneck manifests through multiple independent mechanisms, rather than a single unified process. This observation supports the theoretical framework presented in our introduction, where different aspects of visual information face distinct challenges during cross-modal translation.

\section{Conclusion}
In this work, we provided empirical characterization of the describe-then-generate bottleneck in multimodal AI systems. We generated 150 image pairs through the describe-then-generate pipeline from existing prompts and applied existing evaluation metrics (LPIPS, SSIM, and color distance) to measure information preservation. Our results reveal near-universal perceptual degradation (99.3\% of samples) and substantial structural information loss (91.5\% of samples), providing evidence that the describe-then-generate bottleneck represents a measurable limitation in current multimodal systems. The weak correlations between different degradation dimensions indicate that information loss occurs through multiple independent mechanisms. This characterization provides baseline measurements of cross-modal information loss in vision-language-vision pipelines.

\section{Limitations and Future Work}

This study focuses exclusively on Gemini-2.5-Flash for both vision-language description and image generation. While this controlled approach effectively isolates the bottleneck effect within a single system, it limits the generalizability of our specific quantitative findings across different VLM and generative architectures. Additionally, our evaluation employs a simple "Describe the Image" prompt, representing only one approach to VLM-based description. Future work should systematically evaluate bottleneck effects across multiple architectures to establish the generality of these phenomena and investigate whether enhanced prompting strategies or alternative intermediate representations could mitigate information loss while preserving the efficiency benefits of textual intermediation.
\bibliographystyle{unsrt}  
\bibliography{references}

\end{document}